\documentclass[conference]{IEEEtran}
\IEEEoverridecommandlockouts
\usepackage{cite}
\usepackage{amsmath,amssymb,amsfonts}
\usepackage{algorithmic}
\usepackage{graphicx}
\graphicspath{{image/}}
\usepackage{textcomp}
\usepackage{xcolor}
\usepackage{hyperref}
\usepackage{verbatim}
\usepackage{algorithm}
\usepackage{booktabs}
\usepackage{multirow}

\def\BibTeX{{\rm B\kern-.05em{\sc i\kern-.025em b}\kern-.08em
    T\kern-.1667em\lower.7ex\hbox{E}\kern-.125emX}}
\begin{document}

\title{Spatio-Temporal Latent Graph Structure Learning for Traffic Forecasting
\thanks{This work was supported by the National Key R\&D Program of China (2019YFB2101801), the Sichuan Science and Technology Program (No. 2021YFG0312) and the National Natural Science Foundation of China (No. 62176221).}
}

\author{\IEEEauthorblockN{Jiabin Tang}
\IEEEauthorblockA{\textit{School of Computing and Artificial} \\
\textit{ Intelligence, }\\
\textit{Southwest Jiaotong University}\\
Chengdu, China \\
tjb2019111290@my.swjtu.edu.cn}
\and
\IEEEauthorblockN{Tang Qian}
\IEEEauthorblockA{\textit{SWTJU-Leeds Joint School, } \\
\textit{Southwest Jiaotong University}\\
Chengdu, China \\
qiantang007@my.swjtu.edu.cn}

\and
\IEEEauthorblockN{Shijing Liu}
\IEEEauthorblockA{\textit{SWTJU-Leeds Joint School, } \\
\textit{Southwest Jiaotong University}\\
Chengdu, China \\
shijingLiu@my.swjtu.edu.cn}
\and
\IEEEauthorblockN{Shengdong Du\thanks{\IEEEauthorrefmark{1} Shengdong Du is the corresponding author.}\IEEEauthorrefmark{1}}

\IEEEauthorblockA{\textit{School of Computing and Artificial} \\
\textit{ Intelligence, }\\
\textit{Southwest Jiaotong University}\\
Chengdu, China \\
sddu@swjtu.edu.cn}
\and
\IEEEauthorblockN{Jie Hu}

\IEEEauthorblockA{\textit{School of Computing and Artificial} \\
\textit{ Intelligence, }\\
\textit{Southwest Jiaotong University}\\
Chengdu, China \\
jiehu@swjtu.edu.cn}
\and
\IEEEauthorblockN{Tianrui Li}
\IEEEauthorblockA{\textit{School of Computing and Artificial} \\
\textit{ Intelligence, }\\
\textit{Southwest Jiaotong University}\\
Chengdu, China \\
trli@swjtu.edu.cn}
}

\maketitle

\begin{abstract}
Accurate traffic forecasting, the foundation of intelligent transportation systems (ITS),
 has never been more significant than nowadays due to the prosperity of smart cities and urban computing.
Recently, Graph Neural Network truly outperforms the traditional methods.
Nevertheless, the most conventional GNN-based model works well while given a pre-defined graph structure.
And the existing methods of defining the graph structures focus purely on spatial dependencies and ignore the temporal correlation. 
Besides, the semantics of the static pre-defined graph adjacency applied during the whole training progress is always incomplete, thus overlooking the latent topologies that may fine-tune the model.  
To tackle these challenges, we propose a new traffic forecasting framework--Spatio-Temporal Latent Graph Structure Learning networks (ST-LGSL).
More specifically, the model employs a graph generator based on Multilayer perceptron and K-Nearest Neighbor, which learns the latent graph topological information from the entire data considering both spatial and temporal dynamics.
Furthermore, with the initialization of MLP-kNN based on ground-truth adjacency matrix and similarity metric in kNN, ST-LGSL aggregates the topologies focusing on geography and node similarity. 
Additionally, the generated graphs act as the input of the Spatio-temporal prediction module combined with the Diffusion Graph Convolutions and Gated Temporal Convolutions Networks.
Experimental results on two benchmarking datasets in real world  demonstrate that ST-LGSL outperforms various types of state-of-art baselines. 

\end{abstract}


\section{Introduction}
Owing to the proliferation of urbanization, transportation systems have been under heavy stress and intractable.  
Fortunately, with the pervasive use of the sensors deployed in cities, intelligent transportation systems (ITS) as well as smart city has intrigued the public and highlighted manifold merits. 
Traffic forecasting, which means predicting the future traffic volume from past observed data,
truly plays a pivotal role in smart city efforts and Spatio-Temporal data-mining\cite{yiCityTrafficModelingCitywide2019}\cite{xieUrbanFlowPrediction2020}.
Through forecasting future traffic, it provides related departments references in constructing smart cities from managing modern traffic, reducing the risk of public safety emergencies, etc. 
From another perspective, owning a smooth outing plan based on the information brought by traffic forecast, the living quality of most people could be improved as well. 
To sum up, accurate traffic forecasting is a meaningful and urgent problem.

More recently, deep learning technique, especially Graph Neural Networks (GNN), is widely used in traffic forecasting.
The modeling of spatio-temporal traffic forecasting can be divided into two major portions: extracting temporal features and spatial features.  
To capture temporal dynamics, the traffic dataset can be regarded as a time series among nodes on a graph, consequently, the recurrent neural network (RNN) represented by GRU\cite{li_graph_2017}\cite{li_dynamic_2021} and LSTM\cite{shiConvolutionalLSTMNetwork2015} is commonly utilized to model the complex temporal correlations.
However, so as to solve the problem of the high computational complexity of RNN and the accumulation of errors during iteration, gated temporal convolutions, which often consist of 1-D convolution and gated linear units (GLU), are widely used\cite{yuSpatioTemporalGraphConvolutional2018}, to enable models to be more flexible and lightweight as well as facilitate parallel computation.  
Furthermore, in order to build a larger receptive field with the layer increasing, the dilated causal convolution networks represented by WaveNet\cite{oordWaveNetGenerativeModel2016} are used to extract the temporal correlation of the traffic dataset\cite{wuGraphWaveNetDeep2019}\cite{DynamicMultifacetedSpatiotemporal}\cite{noauthor_connecting_nodate}.  
Meanwhile, with the broader applications and excellent performance of the self-attention mechanism represented by transformer in natural language processing and computer vision, the attempt of self-attention in spatio-temporal data-mining, especially in traffic forecasting, has gradually sprung up\cite{MetaGraphTransformer}.

On the other hand, GNN are adopted to represent the spatial topologies while dealing with the non-euclidean data in the real-world traffic forecasting scenario.
Nevertheless, the graph structure, the premise of GNN methods' tremendous performance, isn't generally available or it could be fragmented.
Hence, there are myriad methods and explorations to deal with this issue. 
The four most commonly employed solutions are as below:  
1) \textbf{Pre-defined:} calculating the geographic distance between pair-wise nodes and then using the threshold function\cite{yuSpatioTemporalGraphConvolutional2018}\cite{li_graph_2017}.  
2) \textbf{Adaptive:} multiplying the learnable node embedding to build the graph structure adaptively\cite{baiAdaptiveGraphConvolutional2020}.
3) \textbf{Dynamic:} using hyper-network for fusing the current and historical information of the nodes\cite{li_dynamic_2021} or tensor decomposition reversely\cite{DynamicMultifacetedSpatiotemporal} to generate the graph structure at each time step dynamically. 
4) \textbf{GAT:} calculating the correlation weights based on GAT between the pair-wise nodes\cite{velickovicGraphAttentionNetworks2018} on the fully connected graph to avoid defining general graph adjacency matrices\cite{zhangSpatialTemporalConvolutionalGraph2020}.

However, the aforementioned manners have their own limitations and drawbacks:
\begin{itemize}
\item The pre-defined approaches with the same static graph adjacency during the whole training process are too intuitive and not able to embrace quite complicated topologies about spatial correlations.
\item The adaptive and dynamic methods always initialize a random tensor which is automatically updated during training and the interpretability of the models is poor.
\item The GAT-based approaches don't take into account not only the correlation between higher-order neighbors but also temporal dynamics.
\end{itemize}

To address the above challenges, inspired by graph structure learning applied across numerous domains\cite{zhuDeepGraphStructure2021a}, we propose a new framework Spatio-Temporal Latent Graph Structure Learning networks (ST-LGSL), for traffic forecasting.
Primarily, in ST-LGSL, we adopt a Latent Graph Structure Learning module (LGSLm) within two different categories of graph generators based on Multilayer perceptron (MLP), called MLP-based generator.
The LGSLm utilizes the entire traffic datasets, which could be regarded as each node having a feature vector at all time steps, as the input to mine the latent graph structure and topological information thus extracting both spatial and temporal features.
Then the graph structure is fed into a spatio-temporal prediction module within Diffusion Graph Convolutions and Gated Temporal Convolutions Networks (TCN).
Besides, to shorten training time and settle in a better local optimum, a \emph{curriculum learning} strategy is employed.

We highlight the major contributions of our work as follows:
\begin{itemize}
\item We propose a novel traffic forecasting framework (ST-LGSL) which mines the latent graph structure by MLP, with the cooperation of the stacked Spatio-Temporal prediction layers, which contain diffusion convolutions layers and temporal convolutions layers, and curriculum learning training strategy.
\item ST-LGSL preserves the latent topologies about both node-wise similarity and geographical information, with the help of MLP-kNN graph generator which contains the similarity function and ground-truth initialization method. 
Compared with the aforementioned constructing methods, ST-LGSL also has stronger interpretability and contains high-order features of neighbors.
\item Our extensive experimental results on two real-world datasets (evaluating both traffic speed and traffic flow) show that ST-LGSL outperforms various types of state-of-art baselines.
And we conduct the model efficiency research especially Latent Graph Structure Learning Module during the forecasting process by visualizing the graph structures.
\end{itemize}
\section{Related Work}
\subsection{Traffic Forecasting}
Existing models are divided into two major components in traffic forecasting: Spatial features modelling and Temporal features modelling.
Initially, the statistical and traditional machine learning methods (e.g., auto-regressive (AR), and auto-regressive moving average (ARMA), auto-regressive integrated moving average (ARIMA)\cite{boxTimeSeriesAnalysis2015}\cite{chenShorttimeTrafficFlow2011}, etc. ) applied in traffic predictions regarding the task as a pure time series forecasting study, thus focusing on the single variant.
Furthermore, with the rising growth of deep learning in various domains like Computer Vision and Natural Language Processing, spatial dependency modelling extract increasing attention.
Different deep learning frameworks (i.e., CNN and GNN) and various spatial topology constructions are closely related.
In the beginning, Zhang et al.\cite{zhangDeepSpatioTemporalResidual} employed grid patterns to represent the traffic network structure so that CNN can be applied to extract the spatial topologies.
Moreover, GNN show more vital representational ability for non-European data, naturally suitable for traffic forecasting.
The key to the GNN is the construction of a graph and the methods of aggregating the features among graph nodes.
The pre-defined methods, which consist of distence-meature\cite{li_graph_2017}\cite{yuSpatioTemporalGraphConvolutional2018} and similarity-meature\cite{bai2019passenger} function, are commonly used to constructing the graph structure previously.
To explore more complex topological informations, an adaptive approach was employed in \cite{baiAdaptiveGraphConvolutional2020} and \cite{wuGraphWaveNetDeep2019}, while diverse methods are used to generate dynamic graph at every time step in \cite{DynamicMultifacetedSpatiotemporal}, \cite{li_dynamic_2021}, \cite{diaoDynamicSpatialTemporalGraph2019}, \cite{guoDynamicGraphConvolution2020} and \cite{mallickDynamicGraphNeural2020}.
With the advance of the attention mechanism, GAT, which regards the graph as a full-connected graph and uses the weighted attention matrix to denote the correlation, is employed in \cite{velickovicGraphAttentionNetworks2018} and \cite{zhangSpatialTemporalConvolutionalGraph2020}.
As for aggregating the features from neighbours on graphs, the graph convolutions are commonly used, which is divided into main categories as follows:
1) the Chebyshev which ues Chebyshev polynomial\cite{yuSpatioTemporalGraphConvolutional2018}\cite{guoAttentionBasedSpatialTemporal2019}; 
2) diffusion convolutions which use bidirectional random walks\cite{li_graph_2017}\cite{zhang_traffic_2021}; 
3) mix-hop convolutions which contain information propagation step and information selection step\cite{noauthor_connecting_nodate}\cite{li_dynamic_2021}.
As to temporal dependency modeling, the most common approach is to use RNN, e.g., GRU\cite{li_graph_2017}\cite{li_dynamic_2021} and LSTM\cite{duLSTMBasedEncoderDecoder2019}.
To tackle the limitations of RNN, TCN is widely adopted, e.g., \cite{yuSpatioTemporalGraphConvolutional2018}\cite{noauthor_connecting_nodate}\cite{wuGraphWaveNetDeep2019}\cite{guoAttentionBasedSpatialTemporal2019}, etc.
Also, more recently, more applications of self-attention and transformer in traffic forecasting are emerging\cite{MetaGraphTransformer}.
Instead of separating spatial and temporal dynamics in isolation, the exploration of fusing the two dimensions is more in-depth.
In \cite{liSpatialTemporalFusionGraph2020}, spatial and temporal features are fused in a single graph which is provided for GNN training.
\subsection{Latent Graph Structure Learning}
Owing to the expressive power of GNN applied in non-Euclidean data, GNN has earned a brilliant achievement in various domains.
Nevertheless, GNN performs well only when the graph structures are given properly. 
To address the challenges, which the graph structures are not available or too noisy in the real world, Graph Structure Learning comes into being, especially latent graph structure learning as mentioned in \cite{fatemiSLAPSSelfSupervisionImproves}.
Latent Graph Structure Learning primarily is studied in node classification.
The sampling from the Bernoulli distribution was used in \cite{franceschiLearningDiscreteStructures2020}, which is learned during the training process for each possible edge, to construct graph structures.
In \cite{chen2019deep}, similarity metric and iterative method adopt to learn the graph structure and graph representation simultaneously.
And the graph auto-encoder (GAE) was employed as a graph generator and a novel data augmentation way on graphs was proposed in \cite{zhao2020data}.
In \cite{yuGraphRevisedConvolutionalNetwork2021}, Graph Convolutions Neural networks (GCN) is applied to calibrate the existing graph structures.
The Multilayer perceptron was applied to construct the spatial topological structures in \cite{fatemiSLAPSSelfSupervisionImproves} and \cite{qasimLearningRepresentationsIrregular2019}.
All methods mentioned above learn from the real data to generate the graph structures and mine the latent topologies.
For more details, please refer to \cite{zhuDeepGraphStructure2021a}.
However, Graph Structure Learning rarely applies to Spatio-Temporal data-mining, especially traffic forecasting.
Earlier, the LDS in \cite{franceschiLearningDiscreteStructures2020} was applied to better solve multivariate time series predictions\cite{shangDiscreteGraphStructure2021}.
In this work, we adopt a similar idea to \cite{shangDiscreteGraphStructure2021} and propose a novel Latent Graph Structure Learning framework, which is more suitable for traffic forecasting.

\section{Methodology}
In this section, we primarily formalize problems and definitions of traffic forecasting and then present detailed descriptions of different parts of the proposed ST-LGSL.
The overall framework of the proposed model is illustrated in Figure~\ref{fig:FigureOne}, 
which shows that the entire data is fed into the Graph Generator (Latent Graph Structure Learning module) as graph node features, later the generated graph structure and the sequential data are fed into stacked Spatio-Temporal layers with Gated TCN, Diffusion Graph Convolutions using Curriculum Learning Strategy.
The more details are as follows:
\begin{figure}[htbp]   
	
	\centering
	
	\includegraphics[width=1\linewidth,scale=1.00]{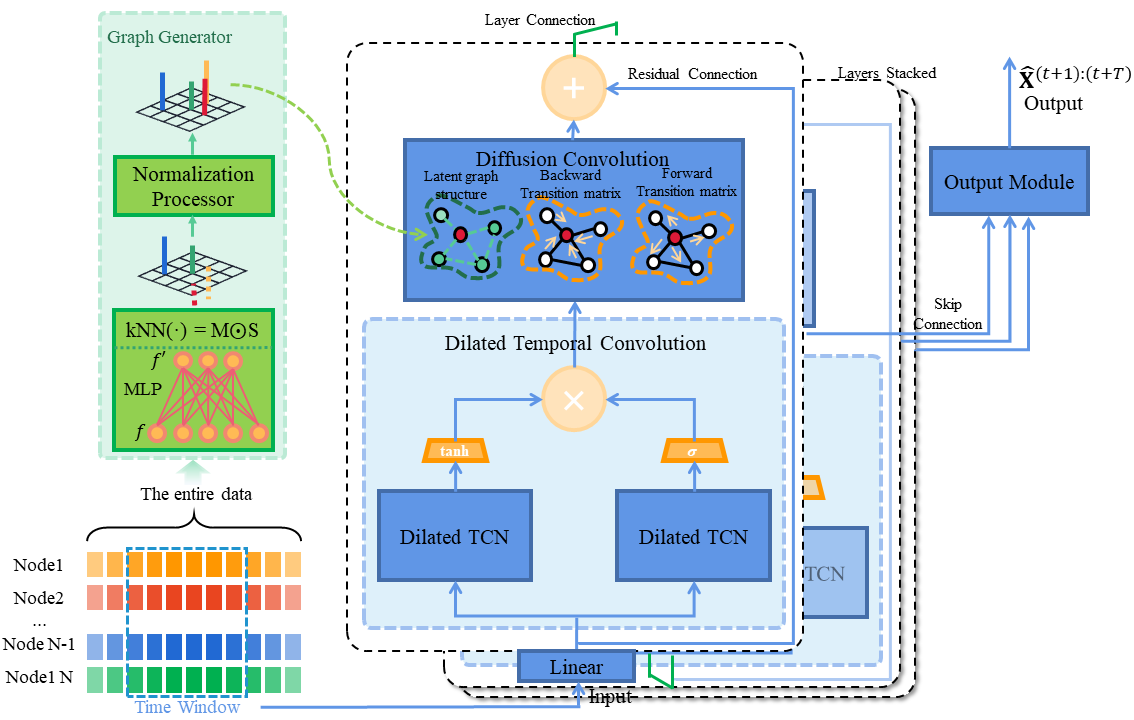}
	
	
	\caption{The framework of the proposed ST-LGSL}
	
	\label{fig:FigureOne}
	
\end{figure}
\subsection{Problems and Definitions}
Traffic forecasting is essentially based on pre-observed traffic data to predict future traffic. 
Generally, we can regard the complex traffic road as a sensor network composed of $M$ sensors, which can be expressed as a weighted directed graph $\mathcal{G}=(V, \varepsilon, A)$. 
Among them, $V$ represents the position of each node of the transportation network, and $|V|=M$; 
$\varepsilon$ is a set of edges and $A \in R^{M \times M}$ is a weighted adjacency matrix used to reflect the proximity between nodes. 
We regard the traffic flow that can be observed in the graph as a graph signal, which can be expressed as $X \in R^{M \times N}$, where $N$ represents the number of features of each node (e.g., traffic volume, speed, etc.).
In each period $t$, we denote $X(t)$ as the observed graph signal. 
In general, the traffic forecasting problem can be summarized as: 
In a given transportation network $\mathcal{G}=(V, \varepsilon, K)$, it could realize the mapping of the graph signal from the historical period $T$ to the future period $T^{\prime}$ after learning a function $p(\cdot)$.
The more details are as follows:

 \begin{equation}
  \begin{array}{c}

    {\left[X_{1+t-T}, \ldots, X_{(t)} ; G\right] \stackrel{p(\cdot)}{\rightarrow}\left[X_{1+t}, \ldots, X_{\left(T^{\prime}+t\right)}\right]} \label{eq13}\\
   
    \\
   
    \text { where, }\left(X_{1+t-T}, \ldots, X_{(t)}\right)=X_{t+1-T: t} \in R^{M \times N \times(T-1)} \\
   
    \left(X_{1+t}, \ldots, X_{\left(T^{\prime}+t\right)}\right)=X_{t+1: t+T^{\prime}} \in R^{M \times N \times\left(T^{\prime}-1\right)}
   
    \end{array}
 \end{equation}

\subsection{Latent Graph Structure Learning}
The Latent Graph Structure Learning module (LGSLm) consists of two components:
1) Graph Generator; 2) Normalization Processor. 
In the rest of this subsection, we initially formalize the general paradigm of LGSL and our two graph generators in detail respectively.
Next, we introduce Normalization Processor more specifically.

\subsubsection{Graph Generator}
Inspired by \cite{fatemiSLAPSSelfSupervisionImproves}, the general paradigm we abide by afterwards of a graph generator is defined as a function Eq.~\eqref{eq1}:
\begin{equation}
	\mathbf{G:}\mathbb{R}^{n\times T}\rightarrow \mathbb{R}^{n\times n}\label{eq1}
\end{equation}
where $n$ denotes the number of the nodes in a traffic dataset and $T$ denotes the number of all recorded time steps.
Hence, the entire data $\mathbf{X}\in \mathbb{R}^{n\times T}$ and the generated graph structure $A^{\prime}\in \mathbb{R}^{n\times n}$.
The function $\mathbf{G}$ contains parameters $\theta_{\mathbf{G:}}$ which tend to learn and optimize during training.
Generally, $\mathbf{G}$ is able to map the serialized traffic dataset to the graph structure in a data-driven way thus learning the latent graph topological information.
However, the direct mapping is hard to train and computationally heavy.
Therefore, we convert Eq.~\eqref{eq1} to Eq.~\eqref{eq2} and propose the following a Graph Generator to overcome it.
\begin{equation}
	\mathbf{G:}\mathbb{R}^{n\times T}\rightarrow \mathbb{R}^{n\times T^{\prime}}\rightarrow \mathbb{R}^{n\times n}, \label{eq2}
\end{equation}
where $T^{\prime}$ denotes the time embedding which is far less than $T$.

\textbf{MLP-kNN:} 
For this generator, we adopt the MLP and k-nearest neighbors to complete the first and second phases of the Eq.~\eqref{eq2} respectively. MLP-kNN is formulized as:
\begin{equation}
	\mathbf{G:} A^{\prime} = kNN(MLP(\mathbf{X}))\label{eq3}
\end{equation}
where $MLP:\mathbb{R}^{n\times T}\rightarrow \mathbb{R}^{n\times T^{\prime}}$ employs MLP to encode the nodes embeddings, $kNN: \mathbb{R}^{n\times T^{\prime}}\rightarrow \mathbb{R}^{n\times n}$ uses kNN to produce a graph adjacency matrix and $\mathbf{X} \in \mathbb{R}^{n\times T}$. 
More specifically, as to the details of kNN, we primarily calculate the \emph{top-k Sim-Matrix} $\mathbf{T} \in\mathbb{R}^{n\times n}$ and \emph{function Sim-Matrix} $\mathbf{S} \in\mathbb{R}^{n\times n}$ which are defined as Eq.~\eqref{eq4}, Eq.~\eqref{eq5}:
\begin{equation}
\mathbf{T}_{ij}=\left\{
\begin{aligned}
1&,&v_j \mbox{ is in the set of top k similar nodes to } v_i \\
0&,&\mbox{otherwise}
\end{aligned}
\right.\label{eq4}
\end{equation}
\begin{equation}
	\mathbf{S}_{ij} = \mathbf{Sim}(E_i, E_j), \label{eq5}
\end{equation}
where $E_i$ and $E_j$ denote the node embeddings of node $i$ and $j$ after passing the MLP, and $\mathbf{Sim}$ is the similarity function (could be cosine, etc.).
Then the $kNN(\cdot)$ is defined as:
\begin{equation}
	A^{\prime}=kNN(E)=\mathbf{T}\odot \mathbf{S}, \label{eq6}
\end{equation}
where the $\odot$ denotes the element-wise Hadamard product and $E$ represents the node embeddings produced by MLP.

To aggregate and enhance the geographical information, we initialize the parameters of MLP-kNN $\theta_{\mathbf{G}}$ before training to enable it to generate the pre-defined adjacency matrix based on the distance function. 
Of course, such an initialization method, when applied for other tasks that do not have a pre-defined graph structure, is not required.

All in all, MLP-kNN generator, which learns the entire temporal features of all nodes and initializes with the ground-truth spatial topologies as the starting point, extracts spatial and temporal dynamics simultaneously.

\subsubsection{Normalization Processor}
$A^{\prime}$ that the graph generator produce may be neither symmetric nor normalized.
To address this issue, let $\tilde{\mathbf{A}}=\mathbf{Sym}(\mathbf{Norm}(A^{\prime}))$.
In particular, we formulize as Eq.~\eqref{eq7}:
\begin{equation}
	\tilde{\mathbf{A}}=\frac{1}{2} \mathbf{D}^{-\frac{1}{2}}\left(\mathrm{ReLU}(A^{\prime})+\mathrm{ReLU}(A^{\prime})^{T}\right) \mathbf{D}^{-\frac{1}{2}}\label{eq7}
\end{equation}	
where $\mathrm{ReLU}$ is element-wise activation function, $\mathbf{D}$ represents the diagonal degree matrices of $\frac{1}{2}(\mathrm{ReLU}(A^{\prime})+\mathrm{ReLU}(A^{\prime})^{T})$.
Further, $\frac{1}{2}(\mathrm{ReLU}(\cdot)+\mathrm{ReLU}(\cdot)^{T})$ corresponds to $\mathbf{Sym}(\cdot)$ and $\frac{1}{2} \mathbf{D}^{-\frac{1}{2}}(\cdot)\mathbf{D}^{-\frac{1}{2}}$ corresponds to $\mathbf{Norm}(\cdot)$. 

\subsection{Gated Temporal Convolutions Layer}
Due to heavy computational complexity and less parallel computation, we replace GRU or LSTM with TCN for our temporal modeling.
In particular, we employ the Gated Temporal Convolutions Layer, which is separated into two portions:
1) \textbf{dilated causal convolution}; 
2) \textbf{Gating mechanisms}. 

\subsubsection{dilated causal convolution}
The receptive field of dilated causal convolution expands exponentially with the increase of the number of layers, while the standard 1-D convolutions enlarge linearly.
Hence, the dilated causal convolution is a special instance of the standard 1-D convolutions.
Formally, the dilated causal convolution is written as:
\begin{equation}
	\mathbf{X} *_d \mathbf{f}(t)=\sum_{i=0}^{K-1} \mathbf{f}(t)_i \mathbf{X}_{t-d \times s}
	\label{eq12}
\end{equation}
where $\mathbf{X} \in \mathbb{R}^T$ and $\mathbf{f}(t)\in \mathbb{R}^K$ represent the feature vector and convolution filters at time step $t$,  respectively, $d$ denotes the dilation factor which actually enables dilated causal convolution to stack fewer layers yet capture increasing features than the standard one.

\subsubsection{Gating mechanisms}
Gating mechanisms are an effective technique widely applied in lots of domains (e.g., LSTM, GRU, etc.).
We use Gating mechanisms combined with the above dilated causal convolution called Gate TCN like \cite{wuGraphWaveNetDeep2019}.
Mathematically, the Gate TCN is formulized as:
\begin{equation}
	\mathbf{h}=tanh\left(\mathbf{\Theta}_{1} *_d \mathcal{X}+\mathbf{b}\right) \odot \sigma\left(\mathbf{\Theta}_{2} *_d \mathcal{X}+\mathbf{c}\right)
\end{equation}
where $tanh(\cdot)$ and $\sigma(\cdot)$ denote the activation functions, $\mathbf{\Theta}_{1}$, $\mathbf{\Theta}_{2}$, $\mathbf{b}$ and $\mathbf{c}$ denote the parameters of convolutions, $*_d$ represents the dilated causal convolution and the $\odot$ denotes the element-wise Hadamard product.

\subsection{Diffusion Convolution Layer}
GCN are broadly adopted to aggregate the spatial features from neighbors, which is a vital operation for GNN.
In this work, we choose the diffusion convolution layer which is proved to be more suitable for spatio-temporal prediction.
Formally, regarding a directed graph, the diffusion convolution layer we employed is written as:
\begin{equation}   
	\mathbf{Z}=\sum_{k=0}^{K} \mathbf{P}_{f}^{k} \mathbf{X} \mathbf{W}_{k 1}+\mathbf{P}_{b}^{k} \mathbf{X} \mathbf{W}_{k 2}+\tilde{\mathbf{A}}^k\mathbf{X} \mathbf{W}_{k 3}, \label{eq8}
\end{equation}
\begin{equation}   
	\mathbf{P}_{f}=\mathbf{A} / rowsum (\mathbf{A}), \label{eq9}
\end{equation}
\begin{equation}   
	\mathbf{P}_{b}=\mathbf{A}^{\mathbf{T}} /rowsum \left(\mathbf{A}^{\mathbf{T}}\right), \label{eq10}
\end{equation}
where $\mathbf{P}_{f}$ and $\mathbf{P}_{b}$ represent the forward transition matrix and backward transition matrix, respectively, $\mathbf{A}$ denotes the ground-truth graph adjacency matrix, $\tilde{\mathbf{A}}$ denotes the latent graph structure which the LGSL module generates, and $\mathbf{W}_{k 1}, \mathbf{W}_{k 2}, \mathbf{W}_{k 3}$ are learnable parameters.
Certainly, the ground-truth graph adjacency matrix is not obligatory, while the ground-truth geographical information is unavailable or incomplete.
Thus, the diffusion convolution layer could be defined as:
\begin{equation}   
	\mathbf{Z}=\sum_{k=0}^{K} \tilde{\mathbf{A}}^k\mathbf{X} \mathbf{W}\label{eq11}
\end{equation}

\subsection{Curriculum Learning Strategy}
In traffic forecasting tasks, it is obvious that long-term predictions are much more difficult than short-term predictions, so long-term predictions contribute more to training loss than short-term predictions. 
Therefore, when designing the loss function, the common attempt is to optimize the overall loss of the model by minimizing the long-term predicted loss.
Inspired by the idea of curriculum learning and \cite{noauthor_connecting_nodate}, we adopt a similar Curriculum Learning Strategy yet without a sub-graph algorithm.
In particular, this training strategy begins with predicting purely next-one step afterward and gradually increases the steps of the predictions until the maximum steps of forecasting (12 steps/1 hour in this task).
	
	
	
	
	
	

Based on the above, we outline the train strategy in Algorithm~\ref{alg3}. 
\begin{algorithm} 
	\caption{Curriculum learning training strategy of ST-LGSL} 
	\label{alg3} 
	\renewcommand{\algorithmicrequire}{\textbf{Input:}}
	\renewcommand{\algorithmicensure}{\textbf{Set:}}
	\begin{algorithmic}
		\REQUIRE The dataset $\mathbf{O}$, node feature $\mathcal{F}$, the model $f(\cdot)$ with the total parameters $\mathbf{\Theta}$, learning rate $\gamma$, batch size $b$, step size $s$
		\ENSURE iterated number $it = 1$, task level $r = 1$
		
		\STATE fetch a batchsize of data ($X \in \mathbb{R}^{b\times T \times N \times D} $, $Y \in\mathbb{R}^{b\times T^{\prime} \times N}$) from $\mathbf{O}$
		\REPEAT
		\IF{$it$ \% $s$ == 0 and $r$ $\leqslant$ $T^{\prime}$  } 
		\STATE $r$ = $r$ + 1
		\ENDIF 
        \STATE compute $\hat{\mathcal{Y}}= f(X[:, :, :, :], \mathcal{F}; \mathbf{\Theta})$ 
		\STATE compute $\mathcal{L} = loss( \hat{\mathcal{Y}}[:, : r, :], Y[:, : r,:])$
		\STATE compute the stochastic gradient of $\mathbf{\Theta}$ according to $\mathcal{L}$.
		\STATE update model parameters $\mathbf{\Theta}$ with their gradients
		
		\STATE $it$ = $it$ + 1
		\UNTIL convergence
	\end{algorithmic} 
\end{algorithm}

\section{Experiment}

In this section, we conduct extensive experiments to evaluate our proposed ST-LGSL model on two real-world datasets.
Moreover, the experimental results are demonstrated compared with various competitive frameworks, both conventional machine learning models and deep learning models.

\subsection{Dataset}
We validate the performance on two public transportation network datasets, e.g. a Los Angeles dataset and PEMS08, which are demonstrated in Table~\ref{tab:addlabel}. 

Both datasets aggregate traffic data in 5 minutes, and each sensor contains 288 data records per day. 
The datasets contain three aspects of information, namely flow, speed, and occupancy.
This work selects traffic speed from PeMSD8 and METR-LA while selecting traffic flow from PeMSD8. 
The spatial adjacency network that may be used for the ground-truth information for each dataset is constructed from the actual road network based on distance, which is similar to the pre-processing in \cite{li_graph_2017}, which is defined as follows:
\begin{equation}
  \mathbf{A}_{v_{i}, v_{j}}=\left\{\begin{array}{l}
    \exp \left(-\frac{d_{v_{i}, v_{j}}^{2}}{\sigma^{2}}\right), \text { if } d_{v_{i}, v_{j}} \leq \kappa \\
    0, \text { otherwise }
    \end{array}\right.
\end{equation} 

\begin{table}[htbp]
  \centering
  \caption{Description of METR-LA and PEMS08}
  \resizebox{0.6\linewidth}{!}{
  \begin{tabular}{lccc}
    \toprule
    Datasets & Nodes & Edges & TimeSteps \\
    \midrule
    METR-LA & 207   & 295374 & 34272 \\
    PEMS08 & 170   & 295   & 17856 \\
    \bottomrule
  \end{tabular}%
  }
  \label{tab:addlabel}%
\end{table}%

\subsection{Baseline Methods}
We compare ST-LGSL with those following models: 
\begin{itemize}
\item \pmb{DCRNN}: DCRNN uses bidirectional random walks and encoder-decoder for the spatial dependency as well as adopts scheduled \cite{li_graph_2017}.
\item \pmb{Graph WaveNet}: GraphWaveNet adds a self-adaptive adjacency matrix into graph convolution and employs wavenet frameworks \cite{wuGraphWaveNetDeep2019}. 
\item \pmb{ASTGCN}: ASTGCN develops a spatial-temporal attention mechanism to capture the spatial- temporal correlations through calculating the attention matrix \cite{guoAttentionBasedSpatialTemporal2019}.
\item \pmb{GMAN}: Graph Multi-Attention Network (GMAN) uses an Encoder-decoder architecture, which consists of multiple attention mechanisms \cite{zhengGMANGraphMultiAttention2020}.
\item \pmb{MTGNN}: MTGNN proposes a novel mix-hop propagation layer and a dilated inception layer for extracting spatial and temporal dynamics \cite{noauthor_connecting_nodate}.
\end{itemize}
\begin{table*}[htbp]
  \centering
  \caption{Performance of speed predictions comparison between baseline models and ST-LGSL on METR-LA and PEMS08 datasets.}
  \resizebox{0.8\linewidth}{!}{
    \begin{tabular}{rlccc|ccc|ccc}
    \toprule
    \toprule
    \multicolumn{1}{l}{Dataset} & Models & \multicolumn{3}{c}{Horizon 3} & \multicolumn{3}{c}{Horizon 6} & \multicolumn{3}{c}{Horizon 12} \\
\cmidrule{3-11}          &       & MAE   & MAPE(\%) & RMSE  & MAE   & MAPE(\%) & RMSE  & MAE   & MAPE(\%) & RMSE \\
    \midrule
    \midrule
    \multicolumn{1}{l}{PEMSD8} & VAR   & 1.1326 & 2.24  & 2.0714 & 1.7166 & 3.56  & 3.3052 & 2.1662 & 4.67  & 4.2502 \\
          & HA    & 2.8118 & 6.31  & 5.6763 & 2.8097 & 6.3   & 5.6731 & 2.8045 & 6.27  & 5.6645 \\
          & DCRNN & 1.1957 & 2.36  & 2.4677 & 1.5349 & 3.25  & \pmb{3.3032} & 1.9051 & 4.27  & 4.1449 \\
          & ASTGCN & 1.3792 & 2.96  & 2.9935 & 1.6445 & 3.58  & 3.6193 & 1.9946 & 4.36  & 4.2879 \\
          & GMAN  & 1.1375 & 2.34  & 2.6752 & \pmb{1.3236} & 2.93  & 3.3952 & 1.5121 & 3.55  & 4.0526 \\
          & Graph WaveNet & 1.1184 & 2.33  & 2.5531 & 1.3847 & 3.2   & 3.5325 & 1.6044 & 3.91  & 4.2426 \\
          & MTGNN & 1.1404 & 2.34  & 2.5736 & 1.3975 & 3.14  & 3.4912 & 1.6324 & 3.94  & 4.2491 \\
\cmidrule{2-11}          & \pmb{ST-LGSL} & \pmb{1.1153} & \pmb{2.19}  & \pmb{2.4654} & 1.3752 & \pmb{2.92}  & 3.3372 & \pmb{1.3333} & \pmb{2.87}  & \pmb{3.1887} \\
    \midrule
    \midrule
    \multicolumn{1}{l}{METR-LA} & VAR   & 4.4250  & 10.2  & 7.8948 & 5.4152 & 12.7  & 9.1347 & 6.5278 & 15.8  & 10.1156 \\
          & HA    & 4.1619  & 13    & 7.8017 & 4.1633 & 13    & 7.8063 & 4.1671 & 13    & 7.8022 \\
          & DCRNN & 2.7720  & 7.3   & 5.3864 & 3.1578 & 8.8   & 6.4517 & 3.6087 & 10.5  & 7.6066 \\
          & ASTGCN & 4.8613  & 9.21  & 9.2724 & 5.4339 & 10.13 & 10.6118 & 6.5139 & 11.64 & 12.5235 \\
          & GMAN  & 2.7716  & 7.25  & 5.4834 & 3.077 & 8.35  & 6.3421 & 3.4068 & \pmb{9.72}  & 7.2201 \\
          & Graph WaveNet & 2.6927  & 6.9   & 5.1537 & 3.0751 & 8.37  & 6.2289 & 3.5364 & 10.01 & 7.3732 \\
          & MTGNN & 2.6906  & 6.86  & 5.1858 & 3.0588 & \pmb{8.19}  & 6.1717 & 3.4989 & 9.87  & 7.2382 \\
\cmidrule{2-11}          & \pmb{ST-LGSL} & \pmb{2.6750}  & \pmb{6.86}  & \pmb{5.0859} & \pmb{3.0537} & 8.4   & \pmb{6.1256} & \pmb{3.5036} & 10.1  & \pmb{7.2195} \\
    \bottomrule
    \bottomrule
    \end{tabular}%
    }
  \label{tab:addlabel1}%
\end{table*}%

\begin{table*}[htbp]
  \centering
  \caption{Performance of flow predictions comparison between baseline models and ST-LGSL on PEMS08 dataset.}
  \resizebox{0.8\linewidth}{!}{
    \begin{tabular}{rl|ccccccc|c}
    \toprule
    \toprule
    \multicolumn{1}{l}{} & Metrics & VAR   & DCRNN & STCCN & ASTGCN & Graph WaveNet & GMAN  & STGCN & \pmb{ST-LGSL} \\
    \midrule
    \midrule
    \multicolumn{1}{l}{Dataset: PeMSD8} & MAE   & 23.4689 & 17.8619 & 18.0223 & 18.6126 & 19.1384 & 14.8743 & 18.0211 & \pmb{14.0690} \\
          & MAPE(\%) & 15.42 & 11.45 & 11.4  & 13.08 & 12.68 & 9.77  & 11.39 & \pmb{9.07} \\
          & RMSE  & 36.3353 & 27.8352 & 27.8326 & 28.1631 & 31.0535 & 24.0675 & 27.8331 & \pmb{22.9468} \\
    \midrule
    \midrule
    \multicolumn{1}{l}{Improvement(\%)} & MAE   & 40.05  & 21.23  & 21.94  & 24.41  & 26.49  & 5.41  & 21.93  & - \\
          & MAPE & 41.18  & 20.79  & 20.44  & 30.66  & 28.47  & 7.16  & 20.37  & - \\
          & RMSE  & 36.85  & 17.56  & 17.55  & 18.52  & 26.11  & 4.66  & 17.56  & - \\
    \bottomrule
    \bottomrule
    \end{tabular}%
    }
  \label{tab:addlabel2}%
\end{table*}%

\subsection{Experiment Settings}
In this experiment, we split the dataset METR-LA and PeMSD8 into training, validation and testing sets with a ratio of 7:2:1. Moreover, we set the batch size as 64.
During training, we adopt \emph{learning rate decay} strategy and choose $1.0 \times 10^{-3}$ as the starting learning rate and weight decay is $0.0001$. 
All experiments run on the Linux OS machine with an NVIDIA GeForce GTX 3080 (12GB) GPU and an NVIDIA Titan \uppercase\expandafter{\romannumeral5} (12GB) GPU. Furthermore, the proposed model is built by Pytorch. 

On ST-LGSL, the arguments are set as follows:  
We set the ground-truth adjacency to double transition and set epoch to 1000 while employing the early stop strategy, where the tolerance is 100.
In the MLP-kNN graph generator, we set the non-linear activation function to ReLU.
As for the initialization of MLP-kNN, we adopt the ground-truth adjacency matrix as the target, the similarity function as cosine, and the $k$ of kNN as 20. In contrast, the epoch of initialization is the hyperparameter that is further discussed in the later subsection.
As to the normalization processor, we choose the symmetric and non-symmetric for comparison.

\subsection{Evaluation Metric}
In our experiments, we choose mean
absolute error (MAE), root mean square error (RMSE) and mean absolute
percentage error (MAPE), which is widely adopted in traffic forecasting tasks, to evaluate our model comprehensively with others. 
The definition of MAE that is similar to \cite{wuGraphWaveNetDeep2019} is as below:
\begin{equation}
  Loss\left(\hat{\mathbf{X}}^{(t+1):(t+T)} ; \boldsymbol{\Theta}\right)=\frac{1}{T\times N} \sum_{i=t+1}^{i=t+T} \sum_{j=1}^{j=N} \left|\hat{\mathbf{X}}_{j }^{(i)}-\mathbf{X}_{j }^{(i)}\right|
\end{equation}
where the $T$ and $N$ are the total number of time steps and graph nodes, respectively. 

\subsection{Experimental Results}

The comparison of different models with ST-LGSL predicting the traffic speed on two different datasets is shown in Table~\ref{tab:addlabel1}, while predicting the traffic flow on PEMSD-08 is illustrated in Table~\ref{tab:addlabel2}.

Experimental results show that our ST-LGSL consistently outperforms the baseline model in speed and flow prediction.

We compare the prediction performance of ST-LGSL and other models in 60 minutes on PeMSD8 and METR-LA datasets from the perspective of speed prediction and focus comprehensively on short-term, mid-term and long-term evaluations.

\begin{table}[htbp]
  \centering
  \caption{Ablation study}
  \resizebox{1.04\linewidth}{!}{
    \begin{tabular}{c|ccccc}
    \toprule
    \toprule
    Methods  & ST-LGSL  & w/o Sym  & \multicolumn{1}{c}{w/o Graph-generator } & w/o Pre-defined & \multicolumn{1}{c}{w/o CL} \\
    \midrule
    \midrule
    MAE   &2.675±0.0036 & 2.6821±0.0004  &  3.0390±0.0226  & 2.7228±0.0192 & 3.0393±0.0083 \\
    MAPE(\%) & 6.88±0.0008 &6.92±0.0009  & 8.35±0.0008 & 6.97±0.0009 & 8.45±0.0008 \\
    RMSE  &  5.0859±0.0386 & 5.1019±0.0113& 6.0382±0.0189 & 5.1956±0.0123 & 6.037±0.0145 \\
    \bottomrule
    \bottomrule
    \end{tabular}}
  \label{tab:addlabel3}%
\end{table}%

\begin{itemize}
\item Our ST-LGSL performs better predictions than the traditional temporal models such as HA and VAR by an overwhelming margin on every dataset because our MLP-kNN generator in the Latent Graph Structure Learning module mines the complex topological information and applies it to the graph convolution so that each node fully aggregates the features of the adjacent nodes according to the learned graph structure.

\item Overall, our ST-LGSL is more significant than all the spatial-temporal models. Compared with DCRNN, ST-LGSL is lower down MAE by 6.72\%, 10.40\%, 30.01\% over the horizons 3, 6 and 9 on the dataset PEMS08; On METR-LA, moreover, compared with GraphWaveNet, ST-LGSL lower down MAE by 0.66\%, 0.70\%, 0.93\%.
This is because unlike DCRNN and MTGNN, which use a static adjacency matrix and ASTGCN, GMAN and Graph WaveNet, which employ an adaptive adjacency matrix based on simple tensor multiplication or attention mechanism, we use MLP-kNN. 
The MLP-kNN module combines geographic information and node similarity relationships to generate latent graph structures, and continuously optimize and tend to converge during training. 

\item In terms of long-term time series forecasting, as the forecasting time length increases, the advantages of our ST-LGSL over GraphWaveNet are becoming increasingly apparent, which embodiments the benefits of Latent Graph Structure Learning module over adaptive graph generator in long-term forecasting.

\item Compared with GMAN in long-term forecasting, our ST-LGSL is slightly inferior in the long-term forecast of the next 1 hour due to the more substantial power of multi-head self-attention in extracting the long-lasting temporal dependency.

\end{itemize}

On the other hand, our ST-LGSL also excels in traffic flow prediction. 

\begin{itemize}

\item ST-LGSL make a significant improvement in traffic flow forecasting. Compared with DCRNN, ST-LGSL lower MAE by 21.23\%, RMSE by 17.56\% and MAPE by 20.79\%; For another spatial-temporal model, Graph WaveNet, MAE, RMSE and MAPE has decreased 26.49\%, 26.11\% and 28.47\%. 
As a latent variable, traffic flow is more closely related to the topology of the road network than traffic speed. Hence, it is easier to mine better latent graph structures to optimize graph convolutional operations.

\end{itemize}

\subsubsection{Ablation Study}
In order to validate the effectiveness of significant components that contribute to the improvement of our ST-LGSL, we conduct an ablation study on METR-LA. We name ST-LGSL without different components as follows:
\begin{itemize}\setlength{\itemsep}{3pt }
  \item \pmb{w/o Sym}: ST-LGSL without symmetrizing the graph.
  \item \pmb{w/o Graph-generator}: ST-LGSL without MLP-kNN graph generator.
  \item \pmb{w/o Pre-defined}: ST-LGSL without a pre-difined graph as the ground-truth during the initialization of MLP-kNN.
  \item \pmb{w/o CL}: ST-LGSL without curriculum learning, which means training without gradually increasing the length of prediction.
 
\end{itemize}

\begin{figure}[htbp]
    \centering

    \includegraphics[width=0.5\textwidth]{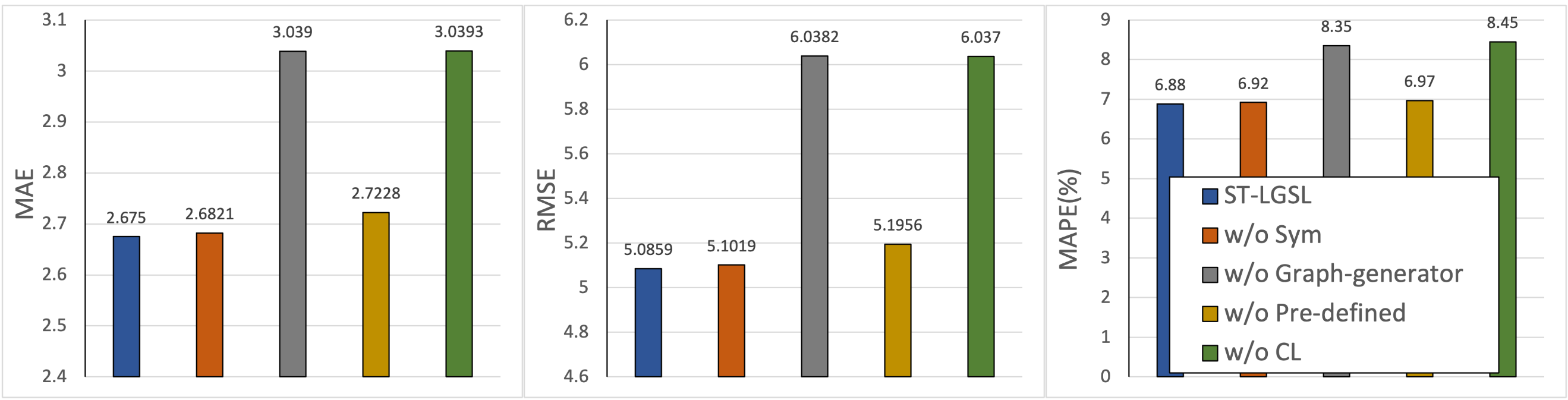}
    \caption{Ablation Study}
    \label{fig:FigureFour}
\end{figure}

We repeat above different models ten times with the same experimental settings as before, and the results are shown in Table~\ref{tab:addlabel3} and Figure~\ref{fig:FigureFour}.
Regarding pre-defined as the initialized target significantly improves the performance because the MLP-kNN obtains the geographical information at the beginning of the training. 
The operations of symmetrizing are practical because the real-word adjacency is symmetric. 
Similarly, curriculum learning is effective, making our model converge quickly and settle in a better local optimum. 
The effect of the graph generator is highly significant 
by making full use of the entire temporal features of all nodes.

\begin{table}[htbp]
  \centering
  \caption{Hyperparameter}
  \resizebox{\linewidth}{!}{
    \begin{tabular}{l|cccc}
    \toprule
    \toprule
    Epoch & 10    & 100   & 1000  & 10000 \\
    \midrule
    \midrule
    MAE   & 2.6833±0.0098 &  2.6913±0.0128 & 2.6752±0.0004  & 2.6889±0.0039 \\
    MAPE(\%) & 6.97±0.0008  &  6.9±0.0009   &  6.88±0.0009 & 6.93±0.0009 \\
    RMSE  & 5.1314±0.0413  & 5.1307±0.0329  & 5.0859±0.0386   & 5.134±0.0233 \\
    \bottomrule
    \bottomrule
    \end{tabular}}%
  \label{tab:addlabel4}%
\end{table}%

\subsubsection{Hyperparameter}
To improve the performance of ST-LGSL, we also carry out additional work on the hyperparameter set: The number of Epoch while initializing the parameters in the MLP-kNN graph generator. 
MLP-epoch means when all the data is fed into the MLP-kNN graph generator, the number of forwarding calculations and backpropagation times. 
We repeat the different cases over 15 times and calculate the average of MAE, RMSE, and MAPE in Table~\ref{tab:addlabel4}.
Extensive experimental results demonstrate that when the mlp-epoch is equal to 1000, the model's performance is the best, and the performance will be degraded if it is too large or too small. 

\begin{figure}[htbp]
    \centering
    
    \includegraphics[width=0.5\textwidth]{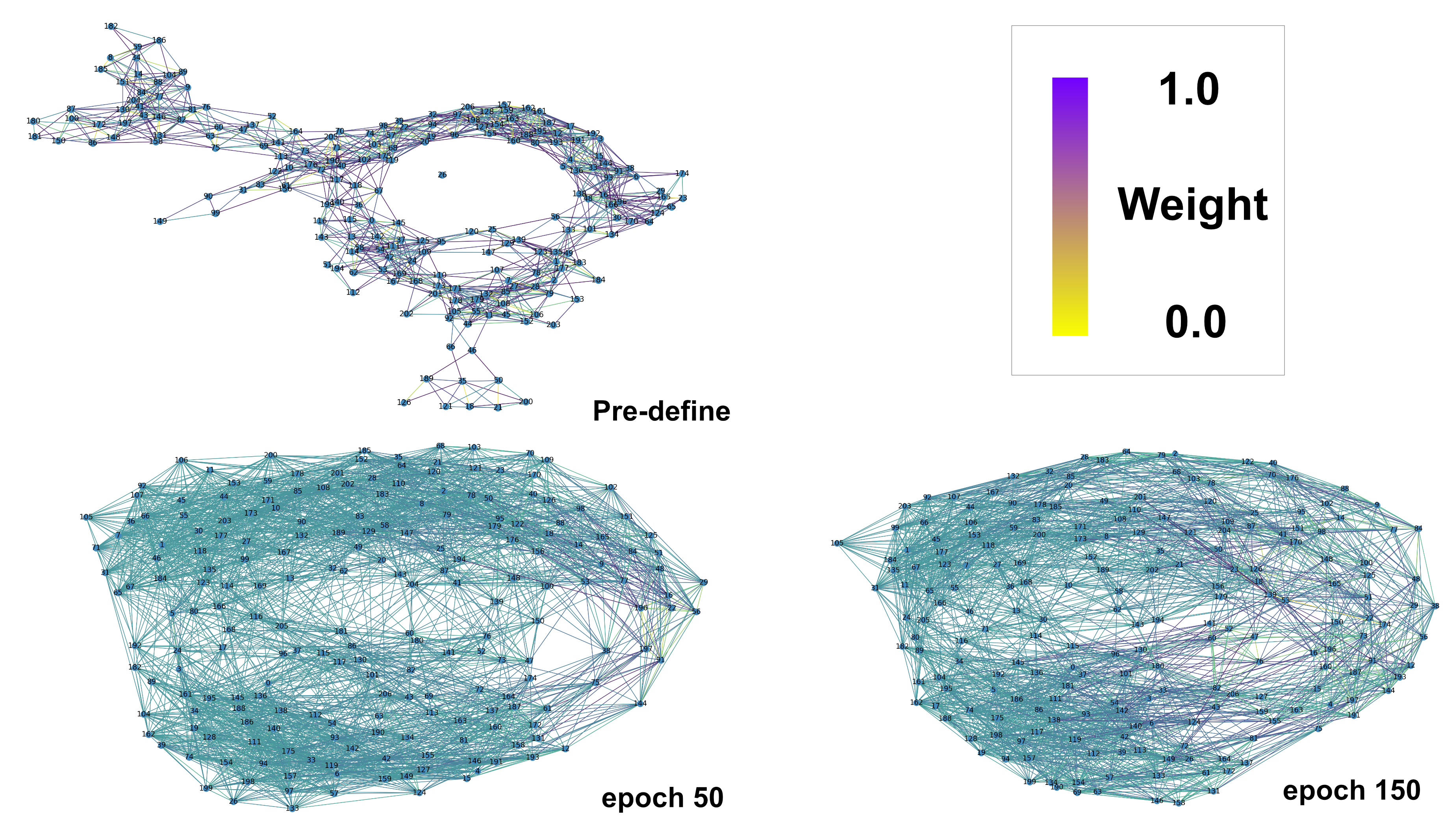}
    \caption{Adjacency Matrix Visualization}
    \label{fig:FigureThree}
\end{figure}

\subsubsection{Learned Graph Structure Analysis}
In order to analyse the Learned Graph Structure intuitively, we visualize the adjacency matrix on the METR-LA dataset in Figure~\ref{fig:FigureThree}, which contains the adjacency matrix in a pre-defined way using the distance function as mentioned above, and epochs 50 and 150 during training, respectively. 
The 207 nodes represent the 207 sensors in the METR-LA dataset, the connection of the two nodes represents the spatial proximity between the two sensors, and the weight of the adjacency matrix is represented by the colour of the edges as the weight increases from 0 to 1. The colour of the edge will gradually deepen. Figure~\ref{fig:FigureThree} can reflect as the model is trained, adjacency is dynamically learning and updating so that the latent graph topologies are learned.
Compared with the pre-defined graph structure, the latent graph structures obtain more node-wise correlation as the train epoch increases.

\section{conclusion}
In this work, we proposed a framework, Spatio-Temporal Latent Graph Structure Learning networks (ST-LGSL), for traffic forecasting.
Our ST-LGSL was equipped with a Latent Graph Structure Learning module (LGSLm), which learns from the entire dataset as graph nodes with time-aware features to extract spatial and temporal dependency.
The LGSLm consists of an MLP-kNN graph generator that employs the ground-truth initialization and similarity metric to obtain geographical and node-similarity information in the learned graph structure.
Moreover, the latent graph structure was fed into the Spatio-temporal predicting framework, which contains the gated temporal convolution and diffusion convolutions layers.
The experimental results demonstrated that our ST-LGSL significantly outperforms various baselines.
In the future, we tend to add self-attention into temporal features modeling to solve the long-term forecasting better.



\bibliography{iclr2022_conference}

\begin{thebibliography}{10}
\providecommand{\url}[1]{#1}
\csname url@samestyle\endcsname
\providecommand{\newblock}{\relax}
\providecommand{\bibinfo}[2]{#2}
\providecommand{\BIBentrySTDinterwordspacing}{\spaceskip=0pt\relax}
\providecommand{\BIBentryALTinterwordstretchfactor}{4}
\providecommand{\BIBentryALTinterwordspacing}{\spaceskip=\fontdimen2\font plus
\BIBentryALTinterwordstretchfactor\fontdimen3\font minus
  \fontdimen4\font\relax}
\providecommand{\BIBforeignlanguage}[2]{{%
\expandafter\ifx\csname l@#1\endcsname\relax
\typeout{** WARNING: IEEEtran.bst: No hyphenation pattern has been}%
\typeout{** loaded for the language `#1'. Using the pattern for}%
\typeout{** the default language instead.}%
\else
\language=\csname l@#1\endcsname
\fi
#2}}
\providecommand{\BIBdecl}{\relax}
\BIBdecl

\bibitem{yiCityTrafficModelingCitywide2019}
X.~Yi, Z.~Duan, T.~Li, T.~Li, J.~Zhang, and Y.~Zheng, ``{{CityTraffic}}:
  {{Modeling Citywide Traffic}} via {{Neural Memorization}} and
  {{Generalization Approach}},'' in \emph{Proceedings of the 28th {{ACM
  International Conference}} on {{Information}} and {{Knowledge Management}}},
  2019, pp. 2665--2671.

\bibitem{xieUrbanFlowPrediction2020}
P.~Xie, T.~Li, J.~Liu, S.~Du, X.~Yang, and J.~Zhang, ``Urban flow prediction
  from spatiotemporal data using machine learning: {{A}} survey,''
  \emph{Information Fusion}, vol.~59, pp. 1--12, 2020.

\bibitem{li_graph_2017}
Y.~Li, R.~Yu, C.~Shahabi, and Y.~Liu, ``Diffusion convolutional recurrent
  neural network: Data-driven traffic forecasting,'' in \emph{International
  Conference on Learning Representations}, 2018.

\bibitem{li_dynamic_2021}
F.~Li, J.~Feng, H.~Yan, G.~Jin, D.~Jin, and Y.~Li, ``Dynamic graph
  convolutional recurrent network for traffic prediction: Benchmark and
  solution,'' \emph{ArXiv}, vol. abs/2104.14917, 2021.

\bibitem{shiConvolutionalLSTMNetwork2015}
X.~Shi, Z.~Chen, H.~Wang, D.-Y. Yeung, W.-K. Wong, and W.-c. Woo,
  ``Convolutional lstm network: A machine learning approach for precipitation
  nowcasting,'' in \emph{Advances in neural information processing systems},
  2015, pp. 802--810.

\bibitem{yuSpatioTemporalGraphConvolutional2018}
B.~Yu, H.~Yin, and Z.~Zhu, ``Spatio-temporal graph convolutional networks: A
  deep learning framework for traffic forecasting,'' in \emph{Proceedings of
  the 27th International Joint Conference on Artificial Intelligence}, 2017,
  pp. 3634--3640.

\bibitem{oordWaveNetGenerativeModel2016}
A.~v.~d. Oord, S.~Dieleman, H.~Zen, K.~Simonyan, O.~Vinyals, A.~Graves,
  N.~Kalchbrenner, A.~Senior, and K.~Kavukcuoglu, ``Wavenet: A generative model
  for raw audio,'' \emph{arXiv preprint arXiv:1609.03499}, 2016.

\bibitem{wuGraphWaveNetDeep2019}
Z.~Wu, S.~Pan, G.~Long, J.~Jiang, and C.~Zhang, ``Graph {WaveNet} for deep
  spatial-temporal graph modeling,'' in \emph{Proceedings of the twenty-eighth
  international joint conference on artificial intelligence, {IJCAI}-19}, 2019,
  pp. 1907--1913.

\bibitem{DynamicMultifacetedSpatiotemporal}
L.~Han, B.~Du, L.~Sun, Y.~Fu, Y.~Lv, and H.~Xiong, ``Dynamic and multi-faceted
  spatio-temporal deep learning for traffic speed forecasting,'' in
  \emph{Proceedings of the 27th ACM SIGKDD Conference on Knowledge Discovery \&
  Data Mining}, 2021, pp. 547--555.

\bibitem{noauthor_connecting_nodate}
Z.~Wu, S.~Pan, G.~Long, J.~Jiang, X.~Chang, and C.~Zhang, ``Connecting the
  dots: Multivariate time series forecasting with graph neural networks,'' in
  \emph{Proceedings of the 26th ACM SIGKDD International Conference on
  Knowledge Discovery \& Data Mining}, 2020, pp. 753--763.

\bibitem{MetaGraphTransformer}
X.~Ye, S.~Fang, F.~Sun, C.~Zhang, and S.~Xiang, ``Meta graph transformer: A
  novel framework for spatial-temporal traffic prediction,''
  \emph{Neurocomputing}, 2021.

\bibitem{baiAdaptiveGraphConvolutional2020}
L.~Bai, L.~Yao, C.~Li, X.~Wang, and C.~Wang, ``Adaptive graph convolutional
  recurrent network for traffic forecasting,'' in \emph{Advances in neural
  information processing systems}, 2020, pp. 17\,804--17\,815.

\bibitem{velickovicGraphAttentionNetworks2018}
P.~Veličković, G.~Cucurull, A.~Casanova, A.~Romero, P.~Liò, and Y.~Bengio,
  ``Graph attention networks,'' in \emph{International Conference on Learning
  Representations}, 2018.

\bibitem{zhangSpatialTemporalConvolutionalGraph2020}
X.~Zhang, C.~Huang, Y.~Xu, and L.~Xia, ``Spatial-{{Temporal Convolutional Graph
  Attention Networks}} for {{Citywide Traffic Flow Forecasting}},'' in
  \emph{Proceedings of the 29th {{ACM International Conference}} on
  {{Information}} \& {{Knowledge Management}}}, 2020, pp. 1853--1862.

\bibitem{zhuDeepGraphStructure2021a}
Y.~Zhu, W.~Xu, J.~Zhang, Q.~Liu, S.~Wu, and L.~Wang, ``Deep graph structure
  learning for robust representations: A survey,'' \emph{ArXiv}, vol.
  abs/2103.03036, 2021.

\bibitem{boxTimeSeriesAnalysis2015}
G.~E.~P. Box, G.~M. Jenkins, G.~C. Reinsel, and G.~M. Ljung, \emph{Time
  {{Series Analysis}}: {{Forecasting}} and {{Control}}}.\hskip 1em plus 0.5em
  minus 0.4em\relax {John Wiley \& Sons}, 2015.

\bibitem{chenShorttimeTrafficFlow2011}
C.~Chen, J.~Hu, Q.~Meng, and Y.~Zhang, ``Short-time traffic flow prediction
  with arima-garch model,'' in \emph{2011 IEEE Intelligent Vehicles Symposium
  (IV)}, 2011, pp. 607--612.

\bibitem{zhangDeepSpatioTemporalResidual}
J.~Zhang, Y.~Zheng, and D.~Qi, ``Deep spatio-temporal residual networks for
  citywide crowd flows prediction,'' in \emph{Thirty-first AAAI conference on
  artificial intelligence}, 2017.

\bibitem{bai2019passenger}
L.~Bai, L.~Yao, S.~S. Kanhere, Z.~Yang, J.~Chu, and X.~Wang, ``Passenger demand
  forecasting with multi-task convolutional recurrent neural networks,'' in
  \emph{Advances in {Knowledge} {Discovery} and {Data} {Mining}}, 2019, pp.
  29--42.

\bibitem{diaoDynamicSpatialTemporalGraph2019}
Z.~Diao, X.~Wang, D.~Zhang, Y.~Liu, K.~Xie, and S.~He, ``Dynamic
  spatial-temporal graph convolutional neural networks for traffic
  forecasting,'' in \emph{Proceedings of the AAAI conference on artificial
  intelligence}, 2019, pp. 890--897.

\bibitem{guoDynamicGraphConvolution2020}
K.~Guo, Y.~Hu, Z.~Qian, Y.~Sun, J.~Gao, and B.~Yin, ``Dynamic graph convolution
  network for traffic forecasting based on latent network of laplace matrix
  estimation,'' \emph{IEEE Transactions on Intelligent Transportation Systems},
  pp. 1--10, 2020.

\bibitem{mallickDynamicGraphNeural2020}
T.~Mallick, M.~Kiran, B.~Mohammed, and P.~Balaprakash, ``Dynamic graph neural
  network for traffic forecasting in wide area networks,'' in \emph{2020 IEEE
  International Conference on Big Data (Big Data)}, 2020, pp. 1--10.

\bibitem{guoAttentionBasedSpatialTemporal2019}
S.~Guo, Y.~Lin, N.~Feng, C.~Song, and H.~Wan, ``Attention based
  spatial-temporal graph convolutional networks for traffic flow forecasting,''
  in \emph{Proceedings of the AAAI Conference on Artificial Intelligence},
  2019, pp. 922--929.

\bibitem{zhang_traffic_2021}
X.~Zhang, C.~Huang, Y.~Xu, L.~Xia, P.~Dai, L.~Bo, J.~Zhang, and Y.~Zheng,
  ``Traffic {Flow} {Forecasting} with {Spatial}-{Temporal} {Graph} {Diffusion}
  {Network},'' in \emph{Proceedings of the AAAI Conference on Artificial
  Intelligence}, 2020, pp. 15\,008--15\,015.

\bibitem{duLSTMBasedEncoderDecoder2019}
S.~Du, T.~Li, Y.~Yang, X.~Gong, and S.-J. Horng, ``An lstm based
  encoder-decoder model for multistep traffic flow prediction,'' in \emph{2019
  International Joint Conference on Neural Networks (IJCNN)}, 2019, pp. 1--8.

\bibitem{liSpatialTemporalFusionGraph2020}
M.~Li and Z.~Zhu, ``Spatial-temporal fusion graph neural networks for traffic
  flow forecasting,'' in \emph{Proceedings of the AAAI Conference on Artificial
  Intelligence}, 2020, pp. 4189--4196.

\bibitem{fatemiSLAPSSelfSupervisionImproves}
B.~Fatemi, L.~E. Asri, and S.~M. Kazemi, ``{SLAPS}: Self-supervision improves
  structure learning for graph neural networks,'' in \emph{Advances in Neural
  Information Processing Systems}, 2021.

\bibitem{franceschiLearningDiscreteStructures2020}
L.~Franceschi, M.~Niepert, M.~Pontil, and X.~He, ``Learning discrete structures
  for graph neural networks,'' in \emph{International conference on machine
  learning}.\hskip 1em plus 0.5em minus 0.4em\relax PMLR, 2019, pp. 1972--1982.

\bibitem{chen2019deep}
Y.~Chen, L.~Wu, and M.~J. Zaki, ``Deep iterative and adaptive learning for
  graph neural networks,'' \emph{arXiv preprint arXiv:1912.07832}, 2019.

\bibitem{zhao2020data}
T.~Zhao, Y.~Liu, L.~Neves, O.~Woodford, M.~Jiang, and N.~Shah, ``Data
  augmentation for graph neural networks,'' in \emph{Proceedings of the AAAI
  Conference on Artificial Intelligence}, 2021, pp. 11\,015--11\,023.

\bibitem{yuGraphRevisedConvolutionalNetwork2021}
D.~Yu, R.~Zhang, Z.~Jiang, Y.~Wu, and Y.~Yang, ``Graph-revised convolutional
  network,'' in \emph{Machine learning and knowledge discovery in databases},
  2021, pp. 378--393.

\bibitem{qasimLearningRepresentationsIrregular2019}
S.~R. Qasim, J.~Kieseler, Y.~Iiyama, and M.~Pierini, ``Learning representations
  of irregular particle-detector geometry with distance-weighted graph
  networks,'' \emph{The European Physical Journal C}, vol.~79, pp. 1--11, 2019.

\bibitem{shangDiscreteGraphStructure2021}
C.~Shang, J.~Chen, and J.~Bi, ``Discrete graph structure learning for
  forecasting multiple time series,'' in \emph{International Conference on
  Learning Representations}, 2021.

\bibitem{zhengGMANGraphMultiAttention2020}
C.~Zheng, X.~Fan, C.~Wang, and J.~Qi, ``Gman: A graph multi-attention network
  for traffic prediction,'' in \emph{Proceedings of the AAAI Conference on
  Artificial Intelligence}, 2020, pp. 1234--1241.

\end{thebibliography}
\bibliographystyle{IEEEtran}

\end{document}